\documentclass[letterpaper, 10 pt, conference]{ieeeconf}  
\IEEEoverridecommandlockouts                             
\usepackage{graphics} 
\usepackage{graphicx}
\usepackage[ruled,vlined]{algorithm2e}
\usepackage{amssymb}
\usepackage{amsmath}

\usepackage{algorithmic}

\title{\LARGE \bf
LiDAR Odometry Methodologies for Autonomous Driving: A Survey}

\author{Nikhil Jonnavithula$^{1}$, Yecheng Lyu$^{1}$ and Ziming Zhang$^{1}$

\thanks{$^{1}$Nikhil Jonnavithula, Yecheng Lyu and Dr. Ziming Zhang are with the Department of Electrical 
and Computer Engineering, Worcester Polytechnic Institute, Worcester, MA 01609, USA. 
\newline Email: \{njonnavithula,ylyu,zzhang15\}@wpi.edu}%

}

\begin{document}
\maketitle

\begin{abstract}

Vehicle odometry is an essential component of an automated driving system as it computes the vehicle's position and orientation. The odometry module has a higher demand and impact in urban areas where the global navigation satellite system (GNSS) signal is weak and noisy. Traditional visual odometry methods suffer from the diverse illumination status and get disparities during pose estimation, which results in significant errors as the error accumulates. Odometry using light detection and ranging (LiDAR) devices has attracted increasing research interest as LiDAR devices are robust to illumination variations. In this survey, we examine the existing LiDAR odometry methods and summarize the pipeline and delineate the several intermediate steps. Additionally, the existing LiDAR odometry methods are categorized by their correspondence type, and their advantages, disadvantages, and correlations are analyzed across-category and within-category in each step. Finally, we compare the accuracy and the running speed among these methodologies evaluated over the KITTI odometry dataset and outline promising future research directions.
\end{abstract}


\section{Introduction}
Vehicle odometry is a crucial component of the vehicle localization module in an automated driving system. In contrast to the GNSS/INS \cite{zhu2018gnss} based localization that requires external signals (\emph{e.g.} GNSS signal), vehicle odometry takes advantage of local sensors' readings to track the vehicle's movement to get a reliable measurement in scenarios where the external signals are blocked or are highly noisy. For instance, in urban canyons, tunnels, and valleys, the GNSS signals are highly noisy due to the multi-path errors, whereas the odometry modules are not affected. Additionally, the odometry methods localize the vehicle in 3-Dimensions that facilitate vehicle localization in multi-level roads, while 2D GNSS/INS systems get easily confused.



Traditionally, vehicle odometry algorithms \cite{tarrio2015realtime}, \cite{lu2015robust}, \cite{zihao2017event}, \cite{wang2017stereo}, \cite{lianos2018vso}, \cite{yang2018deep}, \cite{wang2019recurrent}, \cite{sheng2019unsupervised}, \cite{zou2020learning}, \cite{yang2020d3vo}, \cite{aqel2016review}, are based on camera frames. However, there are several drawbacks in the visual odometry algorithms: (1) the performance is subjected to the variations in illumination, (2) the odometry is estimated in the image coordinates that is not homogeneous to the world coordinates. On the contrary, light detection and ranging (LiDAR) devices \cite{roriz2021automotive} actively emit laser beams to avoid the effects of illumination changes and measure the range directly in the world coordinates, which makes them ideal for vehicle odometry tasks.

In the recent decades, LiDAR-based vehicle odometry has attracted increasing research interests. In this task, a LiDAR frame is formatted as a point cloud in the LiDAR coordinates, and each point represents a scan point on an object. Therefore, the LiDAR frames are formatted as point clouds in different coordinates, and the goal of LiDAR odometry is to estimate the transformation between consecutive LiDAR frames. With those transformations estimated, the pose of any LiDAR frame can be obtained by using homogeneous transformation among the coordinate frames. 

In this paper, we surveyed the existing works in the LiDAR odometry domain. Existing works on point cloud registration are also introduced as they can be adapted to transformation estimation between LiDAR frame pairs, which is a crucial step in LiDAR odometry. The rest of this paper is categorized as follows: In Section \ref{sec:pipeline}, the pipeline of the LiDAR odometry is summarized and divided into four steps: (1) pre-processing, (2) feature extraction, (3) correspondence searching, (4) transformation estimation, and (5) post-processing; In Section \ref{sec:pipeline}, the current works are categorized into three major approaches based on the type of correspondences: (1) point correspondence, (2) distribution correspondence, and (3) network feature correspondence; Section \ref{sec:method} introduces the existing algorithms and compare them in each step by their approaches; In Section \ref{sec:performance-comparison}, we examine those algorithms over the KITTI odometry dataset, and compare the precision and running speed. Section \ref{sec:Conclusion} concludes the paper and outlines the promising future research directions.

\section{Pipeline of LiDAR odometry}
\label{sec:pipeline}

\subsection{Problem setting}
A LiDAR frame is formatted as a point cloud that is represented as a set of 3D points. 
Given two LiDAR frames $P_0$ and $P_1$, and their pose $\mathcal{X}_0$ and $\mathcal{X}_1$ in the world coordinates, a point cloud registration algorithm estimates the transformation ${Tr}_0^1: \mathcal{X}_1 \rightarrow \mathcal{X}_0$. As shown in the Eqn \ref{eqn:odom_registration}, the transformation can be formatted as a 4-by-4 matrix, where $R \in SO(3)$ denotes the 3D rotation matrix and $t \in R^3$ denotes the 3D translation vector.

\begin{equation}
\label{eqn:odom_registration}
  Tr = \left[\begin{array}{ccc|c}
        \multicolumn{3}{c|}{\textbf{R}} &   \textbf{t} \\
      \hline \\[-\normalbaselineskip]
      0 &  0 & 0 & 1 \\
    \end{array}\right] \\
\end{equation}

In extension, given the transformation between LiDAR frame pairs, the transformation of any LiDAR frame ${Tr}_0^t: \mathcal{X}_t \rightarrow \mathcal{X}_0$ can be calculated using Equation \ref{eqn:odom_transofrmation}.

\begin{equation}
\label{eqn:odom_transofrmation}
{Tr}_0^t = {Tr}_0^1\ {Tr}_1^2\ \cdots \ {Tr}_{t-1}^t
\end{equation}

Therefore, the key step of LiDAR odometry task is to estimate the transformation between each consecutive LiDAR frame pairs ${Tr}_{t-1}^t: \mathcal{X}_t \rightarrow \mathcal{X}_{t-1}$. 

\subsection{Pipeline}
According to the existing works \cite{li2020dmlo}, \cite{shan2018lego}, \cite{zheng2020lodonet}, the pipeline of LiDAR odometry can be divided into five stages: (1) pre-processing, (2) feature extraction, (3) correspondence searching, (4) transformation estimation, and (5) post-processing. 

In the pre-processing step, LiDAR point clouds are re-organized, segmented, and filtered for better feature extraction and matching. Typical pre-processing methods are 3D-to-2D projection, semantic segmentation, moving object removal and ground removal.
\begin{figure}[t]
\centering

\includegraphics[width=0.8\linewidth]{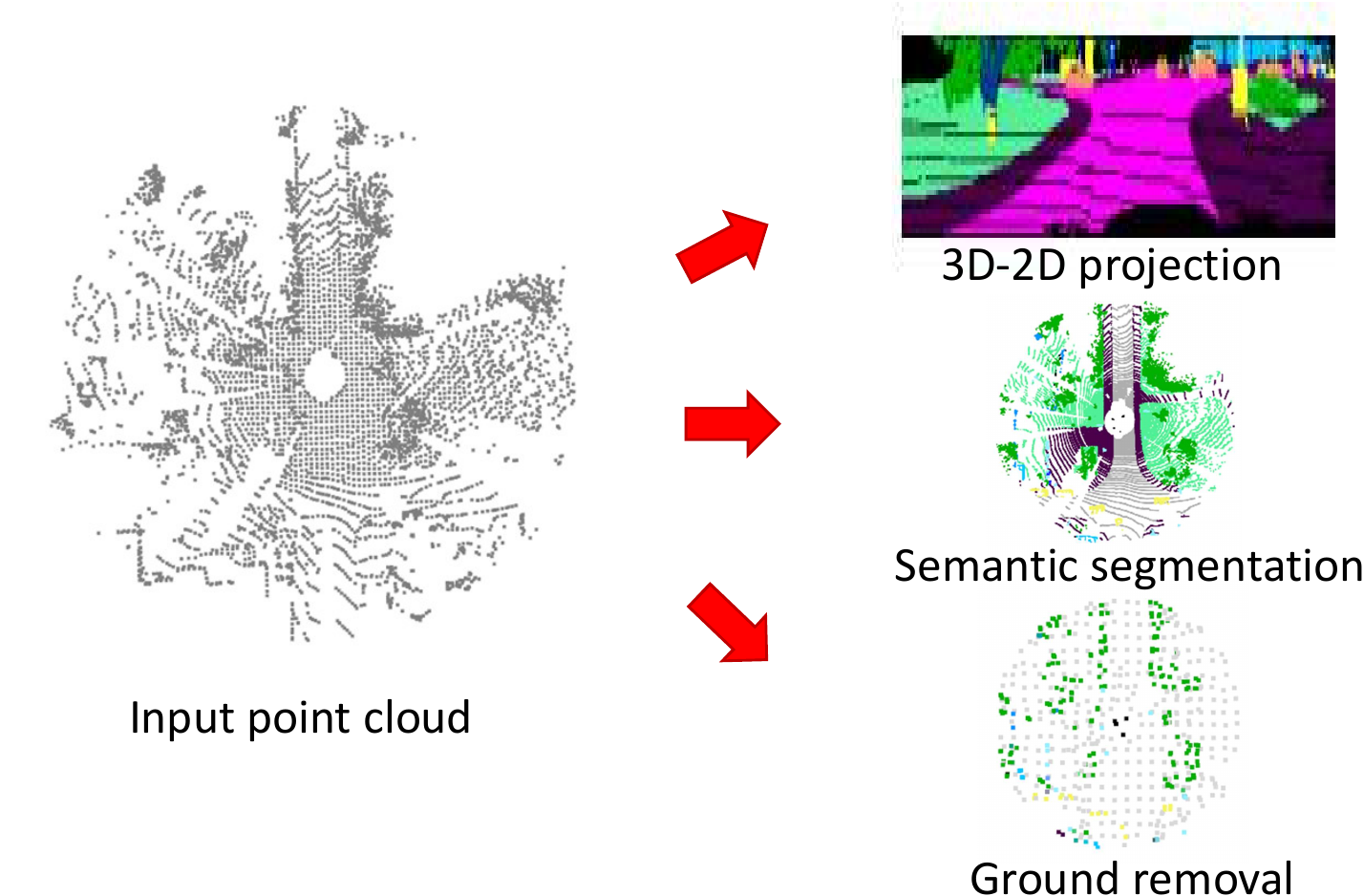}
\caption{An illustration of example tasks in LiDAR frame pre-processing. \label{fig:Preprocessing}}
\vspace{-4mm}
\end{figure}

In the feature extraction step, key points or feature vectors are extracted from points or point clusters, which serve as candidates in feature correspondences. The feature extraction can be achieved by traditional image feature descriptors such as SIFT \cite{lowe2004distinctive}, SURF \cite{bay2006surf}, ORB \cite{rublee2011orb}, or even networks like DCP \cite{wang2019deep}. Other than key points, other features are also utilized, such as normal distribution NDT \cite{biber2003normal} and network embedding as in OverlapNet \cite{chen2021overlapnet}.
\begin{figure}[t]
\centering

\includegraphics[width=\linewidth]{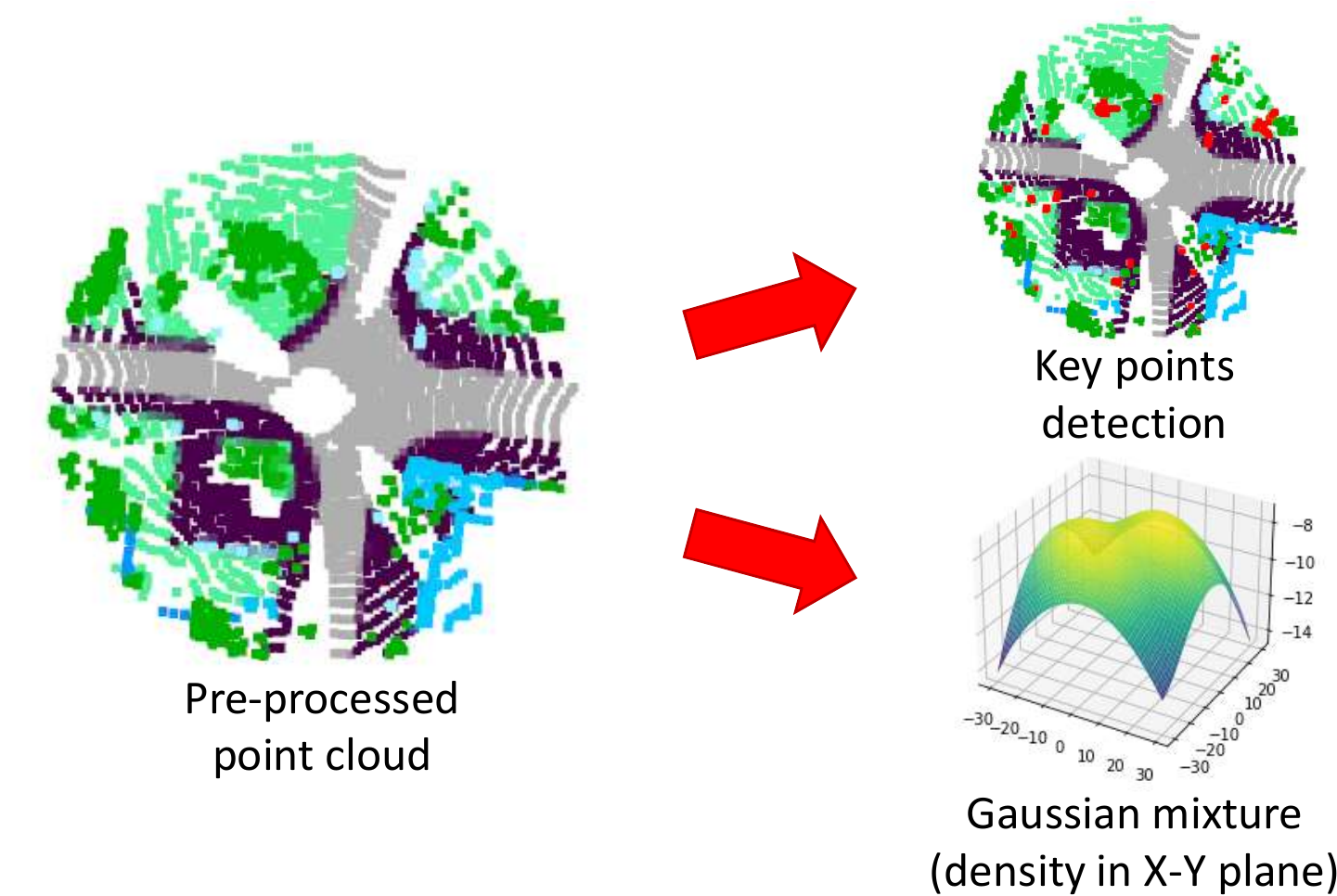}
\caption{An illustration of example tasks in LiDAR frame feature extraction.\label{fig:FeatureExtraction}}
\vspace{-3mm}
\end{figure}

In the correspondence searching step, candidates are matched to generate correspondences. Existing works adopt three types of correspondences: (1) point correspondence, (2) distribution correspondence, and (3) network correspondence. Point correspondence searching is mostly applied since it is straightforward and compatible with existing keypoint matching algorithms. ICP \cite{zhang1992iterative}, RANSAC \cite{fischler1981random}, and neural networks are commonly used point correspondence searching algorithms. Distribution based algorithms do not need correspondence search as the pair of distributions is naturally a correspondence. Similarly, a network feature based algorithm has only one correspondence.

In the transformation estimation step, the methods are highly dependent on the type of correspondences. In general, given a correspondence set $\mathcal{C}{(y_i,x_j)}, y \in \mathcal{X}_{t-1}, x \in \mathcal{X}_t$, and a transformation function $f$,  the goal is to  minimize the loss function as follows:
\begin{equation}
\label{eqn:general_loss_function}
    \min \sum loss (f(\mathcal{X}_{t},\mathcal{X}_{t-1}),y).
\end{equation}

\begin{figure}[t]
\centering

\includegraphics[width=\linewidth]{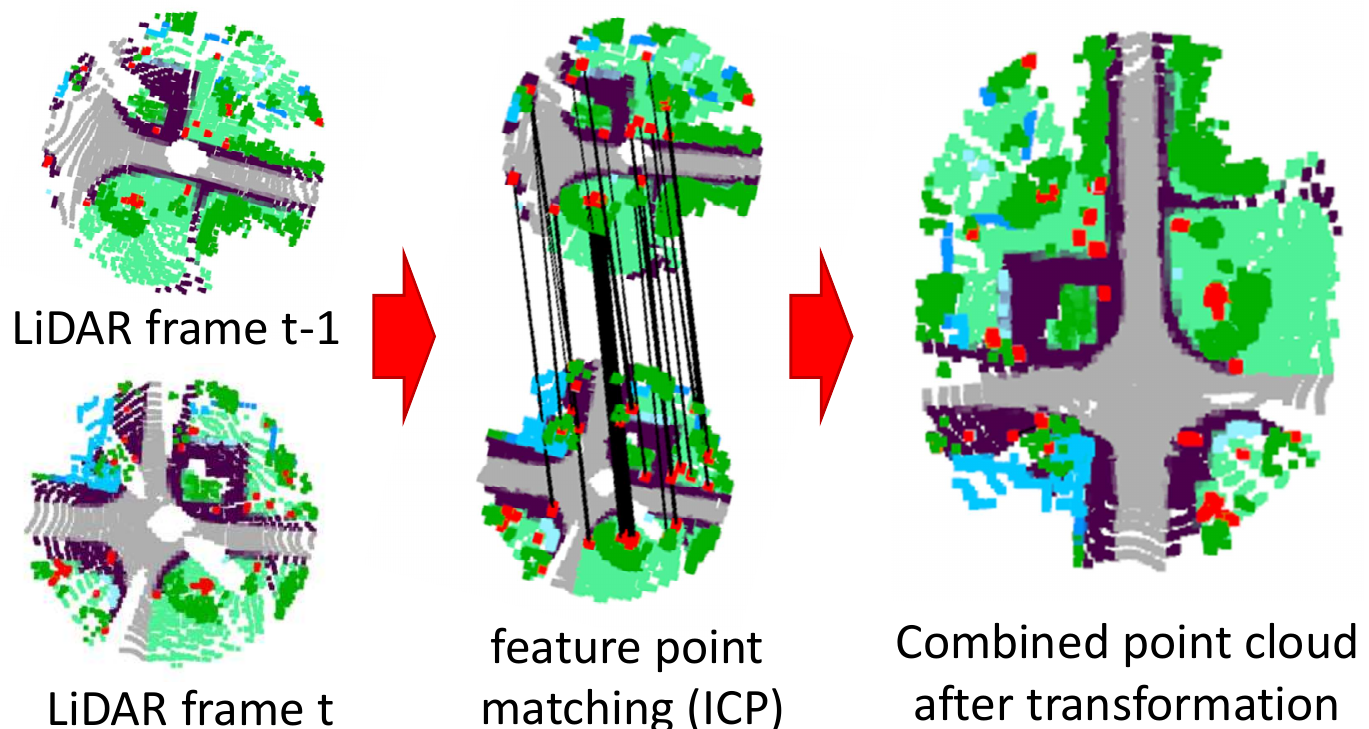}
\caption{An illustration of loop closure, an example task in LiDAR odometry post processing. \label{fig:Transformation}}
\vspace{-3mm}
\end{figure}

In the post-processing step, most existing works calculate the pose of each LiDAR frame using Equation \ref{eqn:odom_transofrmation}. Several works takes advantages of loop closure \cite{chen2021overlapnet}, \cite{wang2020lidar}, \cite{hess2016real} to refine the pose chain \cite{zhang2014loam} \cite{chen2019suma++}.

\begin{figure}[t]
\centering

\includegraphics[width=0.8\linewidth]{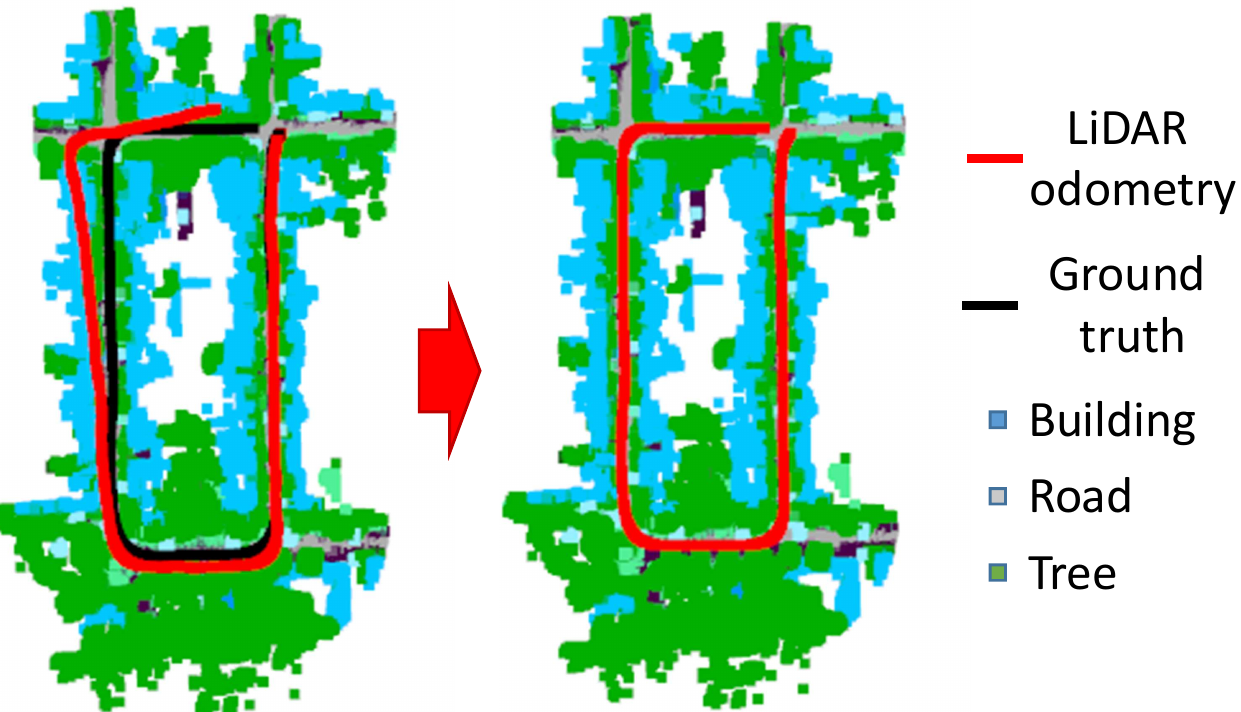}
\caption{An illustration of example tasks in LiDAR frame transformation estimation. \label{fig:Postprocessing}}
\end{figure}

\subsection{loss function for point cloud registration}
For point correspondences, the loss function is 
\begin{equation}
\label{eqn:point_loss_function}
loss = \sum_{(i,j)\in \mathcal{C}} ||y_i-\textbf{R}x_j-\textbf{t}||^2
\end{equation}
, where $y_i$ and $x_j$ are point pairs from the point correspondence $\mathcal{C}$, and $\textbf{R}$ and $\textbf{t}$ are the target transformation parameters. In the existing works,  singular value decomposition (SVD) \cite{kanatani1994analysis} is the most widely used method to to solve the equation.

\begin{figure*}[t]
\centering
\includegraphics[origin=c,width=0.7\linewidth]{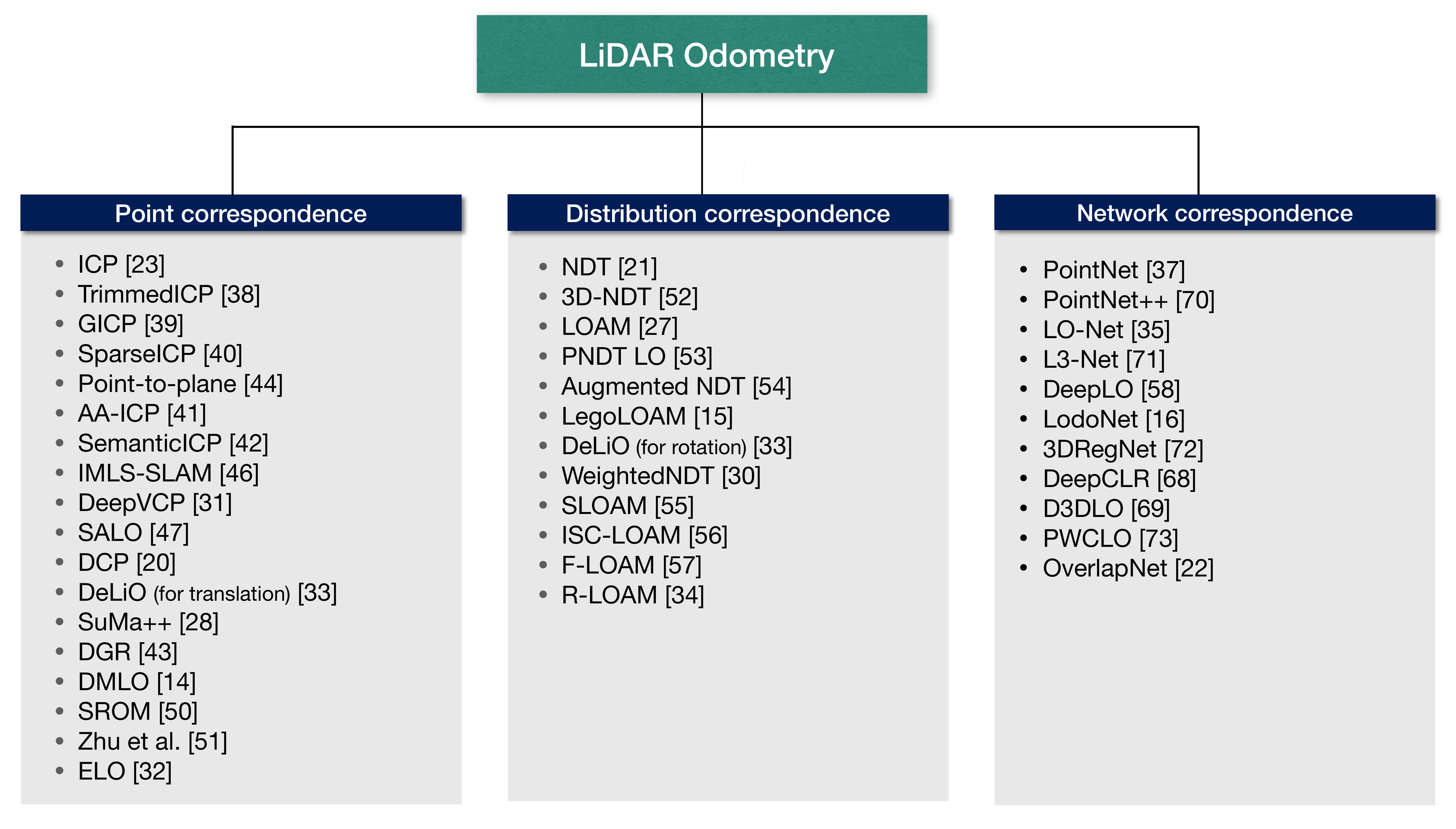}
\caption{Summary of LiDAR odometry methods according to the correspondence type.\label{fig:methods_summary}}
\end{figure*}

For distribution correspondence, normal distribution transform (NDT) \cite{biber2003normal} is the most popular solver. NDT is done by computing the extent of the match with normal distribution instead of the points directly. The normal distribution is obtained by calculating the mean ($\mu$) and covariance matrix ($\Sigma$). As mentioned in \cite{lee2020robust}, under the assumption that a set of point-clouds contained in a voxel is $Y = \{y_1, ..., y_m\}$, the mean and covariance matrix are calculated as follows:
\begin{equation}
\label{eqn:mean}
    \mu=\frac{1}{m} \sum_{k=1}^{m} y_{k}, \quad \Sigma=\frac{1}{m} \sum_{k=1}^{m}\left(y_{k}-\mu\right)\left(y_{k}-\mu\right)^{T}
\end{equation}
The normal distribution of D-dimension is computed as follows, based on ($\mu$) and ($\Sigma$) in Equation \ref{eqn:mean}:
\begin{equation}
\label{eqn:normDist}
    p(x)=\frac{1}{(2 \pi)^{D / 2} \sqrt{|\Sigma|}} \exp \left(-\frac{(x-\mu)^{T} \Sigma^{-1}(x-\mu)}{2}\right)
\end{equation}
 Then, from Equation \ref{eqn:normDist}, the score function of the individual source point is calculated. The target is to maximize the product of the scores of all points. The problem was also modified to calculate the sum of all scores by taking the log-likelihood of the whole expression rather than calculating the product of scores.
\begin{equation}
\label{eqn:score}
    \operatorname{score}(\vec{p})=-\sum_{k=1}^{n} \tilde{p}\left(T\left(\vec{p}, x_{k}\right)\right)
\end{equation}
Here, in Equation \ref{eqn:score},  $T \left(\vec{p}, x_{k}\right)$ is the means to transform $x_{k}$ by vector $\vec{p}$. The scoring sum of all the source points converted by $\vec{p}$ is the final score of $\vec{p}$. An optimization technique based on Newton’s law is employed to find a vector $\vec{p}$ that minimizes the score.
\begin{equation}
\label{eqn:newton}
    H \Delta \vec{p}=-\vec{g}
\end{equation}
Newton’s law is expressed as given in Equation \ref{eqn:newton}, where, $H$ and $\vec{g}$ represent the Hessian matrix and the gradient vector, respectively. They can be obtained as follows:
\begin{equation}
    \begin{aligned}
H_{i j}=& \sum_{k=1}^{n} d_{1} d_{2} \exp \left(-\frac{d_{2}}{2} x_{k}^{\prime} \Sigma_{k}^{-1} x_{k}^{\prime}\right) \\
&\left(-d_{2}\left(x_{k}^{\prime T} \Sigma_{k}^{-1} \frac{\delta x_{k}^{\prime}}{\delta \vec{p}_{i}}\right)\left(x_{k}^{\prime} T_{k}^{-1} \frac{\delta x_{k}^{\prime}}{\delta \vec{p}_{i}}\right)\right.\\
&\left.+x_{k}^{T} \Sigma_{k}^{-1} \frac{\delta^{2} x_{k}^{\prime}}{\delta \vec{p}_{i} \delta \vec{p}_{j}}+\frac{\delta x_{k}^{\prime}}{\delta \vec{p}_{j}} \Sigma_{k}^{-1} \frac{\delta x_{k}^{\prime}}{\delta \vec{p}_{i}}\right)
\end{aligned}
\end{equation}
Here, $x_{k}^{\prime}$ is defined as $T\left(\vec{p}, x_{k}\right) - \mu$. 
\begin{equation}
    g_{i}=\sum_{k=1}^{n} d_{1} d_{2} x_{k}^{\prime T} \Sigma_{k}^{-1} x_{k}^{\prime} \frac{\delta x_{k}^{\prime}}{\delta \vec{p}_{i}} \exp \left(-\frac{d_{2}}{2} x_{k}^{\prime T} \Sigma_{k}^{-1} x_{k}^{\prime}\right)
\end{equation}
The NDT approach computes the Hessian matrix and the gradient vector at each point and calculates $\Delta p$, which minimizes the score by adding all the values. The calculated $\Delta p$ is added to $\vec{p}$, calculated in the previous step, to result in a new $\vec{p}$. Finally, the optimized transformation vector $\vec{p}$ between two point clouds can be obtained by repeating the same process.


Comparing to point correspondence based and distribution correspondence based methods, network correspondence based methods do not use a determined feature extractor to search the pairs. Instead, they use neural networks with millions of parameters to embed a pair of point clouds and then estimate the transformation between them. By training with a huge amount of data samples, methods in this category learn the point cloud patterns in certain scenarios and precisely estimate the transformation between the LiDAR frames.

\section{Method and Algorithm}
\label{sec:method}

Over the past few decades, there have been several methods published to solve the LiDAR odometry problem. Based on the type correspondence used during the point cloud registration step, the works mainly follow three branches: (1) point correspondence based methods (\cite{lu2019deepvcp} \cite{zheng2021efficient} \cite{thomas2019delio} \cite{li2020dmlo}), (2) distribution correspondence based  methods (\cite{oelsch2021r} \cite{zhang2014loam} \cite{shan2018lego}), and (3) network correspondence based methods (\cite{li2019net} \cite{zhou2018voxelnet} \cite{zheng2020lodonet} \cite{qi2017pointnet}). In Section \ref{sec:pipeline}, we introduced the pipeline of a general LiDAR odometry solution, in which the major branches are compared in each step. This is summarized in the Figure \ref{fig:methods_summary}. In this section, the methods in each branch are analyzed and compared at each step.

\subsection{Point correspondence based method}
 One of the earliest approaches, introduced in \cite{zhang1992iterative}, that is still is in use for the LiDAR odometry application is the Iterative Point Cloud (ICP) method. The working of the algorithm is as delineated in Section \ref{sec:pipeline} and Algorithm \ref{alg:icp}. The last three decades have seen a rise in ICP based methods like TrimmedICP \cite{chetverikov2002trimmed}, GICP \cite{segal2009generalized}, SparseICP \cite{bouaziz2013sparse}, AA-ICP \cite{pavlov2018aa}, SemanticICP \cite{parkison2018semantic}, Suma++ \cite{chen2019suma++}, DGR \cite{choy2020deep}, and ELO \cite{zheng2021efficient}.

The point correspondence based methods extract key points from the LiDAR frames, establish the key point pairs, and estimate the transformation according to the pairs.  A primary disadvantage of the ICP method is that when the point clouds are far, it is prone to local minima. TrimmedICP \cite{chetverikov2002trimmed} is an extension of ICP that incorporates the Least Trimmed Squares (LTS) approach enabling the algorithm to function for overlaps under 50\%.

\begin{algorithm}[t]
\SetAlgoLined
\textbf{Input:} \\
\quad \quad Two point-clouds: $A = \{a_i\}, B = \{b_i\} $  \\
\quad \quad An initial transformation: $T_0$ \\
\textbf{Output:}   \\
\quad \quad The correct transformation, $T$, which aligns $A$ \\
\quad \quad and $B$ 
\begin{algorithmic}[1]
    \STATE ${T}$ $\leftarrow$ ${T}_{0}$
    \STATE \textbf{while} not converged \textbf{do}
    \STATE \quad \textbf{for} $i \leftarrow 1$ \textbf{to} $N$ \textbf{do}
    \STATE \quad \quad \quad $m_i \leftarrow$ FindClosestPointInA$(T \cdot b_i)$; 
    \STATE \quad \quad \quad \textbf{if} $||m_i  - T \cdot b_i|| \leq d_{max}$ \textbf{then}
    \STATE \quad \quad \quad \quad $\omega_i \leftarrow 1$;
    \STATE \quad \quad \quad \textbf{else}
    \STATE \quad \quad \quad \quad $\omega_i \leftarrow 0$;
    \STATE \quad \quad \quad \textbf{end if}
    \STATE \quad \textbf{end for}
    \STATE \quad $T \leftarrow$ $\underset{T}{\arg\min}\{ \underset{i}{\sum} \omega_i||T \cdot b_i - m_i||^2 \}$;
    \STATE \textbf{end while}
    
\end{algorithmic}
 \caption{ICP algorithm \cite{segal2009generalized}}
 \label{alg:icp}
\end{algorithm}

Point-to-plane \cite{grant2012point} variant of ICP improves the  performance by taking  advantage  of  surface  normal  information \cite{chen1992object}. Instead of minimizing step 11, Algorithm \ref{alg:icp} the point-to-plane algorithm minimizes  error  along  the  surface  normal  (i.e.  the  projection of $(T*b_{i} - m_{i})$ onto  the  sub-space  spanned  by  the  surface-normal. Generalized ICP \cite{segal2009generalized} is a combination of the ICP and point-to-plane ICP algorithms into a single probabilistic framework. This approach outperforms ICP and the point-to-plane methods and is, in comparison, more robust to incorrect correspondences. GICP, apart from having a similar speed and simplicity to ICP, facilitates the addition of measurement noise, outlier terms, and probabilistic techniques to increase robustness.

Later, approaches like SemanticICP \cite{parkison2018semantic} and IMLS-SLAM \cite{deschaud2018imls} have provided a new direction for tackling the LiDAR odometry problem. IMLS-SLAM pre-processes the dynamic objects and employs a sampling strategy on LiDAR scans to define a model from prior LiDAR sweeps using Implicit Moving Least Squares (IMLS) surface representation. Whereas SemanticICP conducts joint geometric and semantic inference to improve the registration task by incorporating pixelated semantic measurements into estimating the relative transformation between two point clouds. Here, point associations are treated as latent random variables, which leads to an Expectation-Maximization style solution. SemanticICP outperforms GICP \cite{segal2009generalized} due to the amalgamated use of semantic labels and the EM data associations.

Papers like SALO \cite{kovalenko2019sensor}, DCP \cite{wang2019deep}, DeepVCP \cite{lu2019deepvcp} were published alongside the sufel based approaches SuMa and SuMa++. SALO utilizes the LiDAR sensor hardware and improved the ICP algorithm with novel downsampling and point matching rejection methods. Its advantage is the integration of the physics of the sensor for increased precision LiDAR odometry. The DCP paper \cite{wang2019deep} discusses the spherical projection of point cloud data to reduce the dimensionality of the input data. 

Approaches like the SuMa++ \cite{chen2019suma++} have a semantic step, with  point-wise labels provided by RangeNet++ \cite{milioto2019rangenet++}, to filter out the pixels of dynamic objects and add semantic constraints to the scan registration. In doing so, it outperforms SuMa \cite{behley2018efficient}. DGR  \cite{choy2020deep}, DMLO \cite{li2020dmlo}, and SROM \cite{rufus2020srom}, are few recent methods where the three module-based DGR approach incorporates a Procrustes error for odometry estimation.

When the spherical projection is employed, points belonging to different surfaces could be adjacent in the range image. Some of the most recent approaches are ELO \cite{zheng2021efficient}, and Zhu et al.  \cite{zhu2021cylindrical}, with Zhu et al. proposing a different way of handling point cloud sparseness with cylindrical instead of spherical projection and ELO utilizing Bird's eye view as well, tackles the aforementioned problem.
ELO has the smallest runtime among the methods, even outperforming the long-term performer LOAM.

\subsection{Distribution  correspondence based methods}

The Normal Distribution Transform method, introduced in \cite{biber2003normal}, has been a defining type of approach to tackle the registration problem for LiDAR odometry applications. The relevant computations and theory for NDT approaches have been discussed in Section \ref{sec:pipeline}.
Over the last two decades, multiple versions of NDT such as 3DNDT \cite{magnusson20083d}, PNDT-LO \cite{hong2017probabilistic}, AugmentedNDT \cite{akai2017robust} and weightedNDT \cite{lee2020robust} have been proposed. 

\begin{algorithm}[t]
\SetAlgoLined
\textbf{Input:} \\
\quad \quad The source point cloud $X$  \\
\quad \quad The target point cloud $Z$ \\
\quad \quad Initial guess of transformation $\vec{p}_{ini}$ \\
\textbf{Output:}   \\
\quad \quad Final transformation $\vec{p}$ between $X$ and $Z$ \\
\begin{algorithmic}[1]
    \STATE $\vec{p}$ $\leftarrow$ $\vec{p}_{ini}$
    \STATE \textbf{for all} points $z_i$ $\in$ $Z$ \textbf{do}
    \STATE \quad find the cell $Y$ that contains $z_i$
    \STATE \quad classify $z_i$ to entire cells $Y$
    \STATE \textbf{end for}
    \STATE \textbf{for all} cells $Y$ \textbf{do}
    \STATE \quad $Y = \{y_1, ..., y_m\}$
    \STATE \quad $ \mu=\frac{1}{m} \sum_{k=1}^{m} y_{k}, \Sigma=\frac{1}{m} \sum_{k=1}^{m}\left(y_{k}-\mu\right)\left(y_{k}-\mu\right)^{T}$
        
    \STATE \textbf{end for}
    \STATE \textbf{while} not converged \textbf{do}
    \STATE \quad score $\leftarrow$ 0, $\vec{g} \leftarrow$ 0, $H \leftarrow$ 0 
    \STATE \quad \textbf{for all} points $x_i \in X$ \textbf{do} 
    \STATE \quad \quad find the cell $Y$ that contains $T(\vec{p},x_k)$
    \STATE \quad \quad update $\vec{g}, H$
    \STATE \quad \textbf{end for}
    \STATE \quad solve $H \Delta \vec{p} = -\vec{g}$
    \STATE \quad $\vec{p} \leftarrow \vec{p} + \Delta \vec{p} $
    \STATE \textbf{end while}
    
\end{algorithmic}
 \caption{NDT algorithm \cite{lee2020robust}}
 \label{alg:ndt}

\end{algorithm}

The key step in these approaches is to convert the input point clouds into equal cell sizes on which the normal distributions are computed. These are then compared with other normal distributions from other point clouds and are assigned a score. The algorithm finds the rotation and translation that increases the scores. 

The initial NDT method was utilized primarily for 2D scan registrations. 3D-NDT, apart from extending this approach to 3-Dimensions, is advantageous as it forms a smooth piece-wise spatial representation that inturn facilitates a complete 3D-NDT map generation, post registration. This approach also outperforms the intial approaches like the ICP for point cloud registration task \cite{magnusson20083d}.

Next, one of the significant NDT-based approaches for odometry estimation called LOAM \cite{zhang2014loam}, was published that still has a high standing in the   testing performance in the KITTI odometry dataset, Table \ref{tab:KITTI-testing}. It has no pre-processing for odometry.
In LOAM, the feature points on the sharp edges and planar surface patches are selected, smoothness is evaluated \cite{zhang2014loam} to identify edge and plane points and the point-to-edge and point-to-plane scan-matching is employed to arrive at the transformation between two scans. In the PNDT \cite{hong2017probabilistic}, unlike classical NDT, the probability ditribution function of each point is computed while calculating mean and  the covariance, resulting in an improved translational and rotational accuracy. The advantage is that distributions are generated in all the occupied cells irrespective of the resolution. A variant of LOAM called LegoLOAM, introduced in \cite{shan2018lego}, additionally employs label matching to increase the likelihood of finding matches corresponding to the same object between two scans. Also, a two-step LM optimization is incorporated that allows similar accuracy as LOAM, while achieving a 35\% reduced runtime. 

DeLiO \cite{thomas2019delio} for the first time in LiDAR odometry, introduces the decoupled translation and the rotation modules.   But, DeLiO does not deal with dynamic objects. However, an approach called the weightedNDT \cite{lee2020robust} tackles point cloud matching in robust environments by assigning greater or lesser weights to points that have greater or lesser probabilities, respectively \cite{zou2020learning}. A LOAM-variant called SLOAM \cite{chen2020sloam} introduced the idea of semantic features having greater reliability than the texture-based lines and planes. This enables SLOAM to have an advantage in being more robust in unstructured noisy environments. LiDAR based loop closure detection ignores the intensity reading and uses geometric-only descriptor, but, ISC-LOAM \cite{wang2020intensity} leverages the intensity readings to facilitate effective place recognition. 
 Most recent approaches towards LiDAR odometry are F-LOAM \cite{wang2021f} and R-LOAM \cite{oelsch2021r}. F-LOAM outperforms LOAM and LegoLOAM in terms of runtime. Whereas R-LOAM is an improvement upon LOAM as an additional cost for mesh features is incorporated that results in a reduction of median APE, compared to LOAM.

In terms of performance on the KITTI Odometry dataset, LOAM has consistently performed well. Among the LiDAR-only methods available, LOAM has of of the lowest rotation and translational error.




\begin{table*}[]
\caption {Comparision of performances of published methods on LiDAR odometry over KITTI Odometry training data } \label{tab:KITTI-training} 
\resizebox{\textwidth}{!}{\begin{tabular}{c | c | c | c | c | c | c | c | c | c | c | c | c} 
{\bf Methods} & {\bf 00} & {\bf 01} & {\bf 02} & {\bf 03} & {\bf 04} & {\bf 05} & {\bf 06} & {\bf 07} & {\bf 08} & {\bf 09} & {\bf 10} & {\bf Avg}\\ \hline
LOAM \cite{zhang2014loam} & 0.78/0.53 & 1.43/0.55 & 0.92/0.55 & 0.86/0.65 & 0.71/0.50 & 0.57/0.38 & 0.65/0.39 & 0.63/0.50 & 1.12/0.44 & 0.77/0.48 & 0.79/0.57 & 0.85/0.51\\
\hline
ELO \cite{zheng2021efficient} & 0.54/0.20 & 0.61/0.13 & 0.54/0.18 & 0.65/0.27 & 0.32/0.15 & 0.33/0.17 & 0.30/0.13 & 0.31/0.16 & 0.79/0.21 & 0.48/\textbf{0.14} & 0.59/\textbf{0.19} & {0.50/0.18} \\

\hline

IMLS-SLAM  \cite{deschaud2018imls} & -/0.50 & -/0.82 & -/0.53 & -/0.68 & -/0.33 & -/0.32 & -/0.33 & -/0.33 & -/0.80 & -/0.55 & -/0.53 & -/0.55\\
\hline

SUMA++ \cite{chen2019suma++} & \textbf{0.22}/0.64 & 0.46/1.60 & 0.37/1.00 & 0.46/0.67 & 0.26/0.37 & 0.20/0.40 & 0.21/0.46 & 0.19/0.34 & 0.35/1.10 & \textbf{0.23}/0.47 & \textbf{0.28}/0.66 & 0.29/0.70\\
\hline

SALO \cite{kovalenko2019sensor} & 0.91/0.72 & 1.13/0.37 & 0.98/0.45 & 1.76/0.50 & 0.51/0.17 & 0.56/0.29 & 0.48/0.13 & 0.83/0.51 & 1.33/1.43 & 0.64/{0.30} & 0.97/{0.41} & 0.95/0.80 \\ \hline

SuMa \cite{behley2018efficient} & 0.3/0.7 & 0.5/1.7 & 0.4/1.1 & 0.5/0.7 & 0.3/0.4 & 0.2/0.5 & 0.2/0.4 & 0.3/0.4 & 0.4/1.0 & 0.3/0.5 & 0.3/0.7 & 0.3/0.7 \\
\hline

GICP \cite{segal2009generalized} & 1.29/0.64 & 4.39/0.91 & 2.53/0.77 & 1.68/1.08 & 3.76/1.07 & 1.02/0.54 & 0.92/0.46 & 0.64/0.45 & 1.58/0.75 & 1.97/0.77 & 1.31/0.62 & 1.91/0.73 \\ \hline

LO-Net \cite{li2019net} & 1.47/0.72 & 1.36/0.47 & 1.52/0.71 & 1.03/0.66 & 0.51/0.65 & 1.04/0.69 & 0.71/0.50 & 1.70/0.89 & 2.12/0.77 & 1.37/0.58 & 1.80/0.93 & 1.09/0.63\\
\hline
DeepLO \cite{cho2019deeplo} & 0.32/\textbf{0.12} & \textbf{0.16/0.05} & \textbf{0.15/0.05} & \textbf{0.04/0.01} & \textbf{0.01/0.01} & \textbf{0.11/0.07} & \textbf{0.03/0.07} & \textbf{0.08/0.05} & \textbf{0.09/0.04} & 13.35/4.45 & 5.83/3.53 & 1.83/0.76\\
\hline
DeepVCP \cite{lu2019deepvcp} & -/- & -/- & -/- & -/- & -/- & -/- & -/- & -/- & -/- & -/- & -/- & \textbf{0.071/0.164} \\

\hline
\end{tabular}}
\end{table*}

\begin{table}[t]\centering
\caption {Comparision of performances of published methods on LiDAR odometry over KITTI Odometry test data } \label{tab:KITTI-testing} 
\begin{tabular}{c | c | c | c } 
{\bf Published Methods} & {\bf Translation} & {\bf Rotation} & {\bf Runtime}\\ \hline
LOAM \cite{zhang2014loam}  & \textbf{0.55} \% & \textbf{0.0013} [deg/m] & 0.1 s   \\
\hline
MULLS \cite{pan2021mulls}  & 0.65 \% & 0.0019 [deg/m] & 0.08 s  \\
\hline

ELO \cite{zheng2021efficient}  & 0.68 \% & 0.0021 [deg/m] & \textbf{0.005} s  \\
\hline

IMLS-SLAM \cite{deschaud2018imls}  & 0.69 \% & 0.0018 [deg/m] & 1.25 s  \\
\hline

MC2SLAM \cite{neuhaus2018mc2slam}  & 0.69 \% & 0.0016 [deg/m] & 0.1 s  \\
\hline
F-LOAM \cite{oelsch2021r}  & 0.71 \% & 0.0022 [deg/m] & 0.05 s  \\
ISC-LOAM \cite{wang2020intensity}  & 0.72 \% & 0.0022 [deg/m] & 0.1 s  \\
\hline
PSF-LO \cite{chen2020psf}  & 0.82 \% & 0.0032 [deg/m] & 0.2s  \\
\hline

CAE-LO  \cite{yin2020cae}  & 0.86 \% & 0.0025 [deg/m] & 2 s \\
\hline

CPFG-slam \cite{ji2018cpfg}  & 0.87 \% & 0.0025 [deg/m] & 0.03 s  \\
\hline

PNDT LO \cite{hong2017probabilistic}  & 0.89 \% & 0.0030 [deg/m] & 0.2 s \\
\hline
SuMa-MOS \cite{chen2021moving} & 0.99 \% & 0.0033 [deg/m] & 0.1s  \\
\hline
SuMa++ \cite{chen2019suma++}  & 1.06 \% & 0.0034 [deg/m] & 0.1 s  \\
\hline

ULF-ESGVI \cite{yoon2021unsupervised} & 1.07 \% & 0.0036 [deg/m] & 0.3 s  \\
\hline
STEAM-L \cite{tang2019white}  & 1.22 \% & 0.0058 [deg/m] & 0.2 s  \\
\hline

SALO \cite{kovalenko2019sensor}  & 1.37 \% & 0.0051 [deg/m] & 0.6 s \\
\hline

SuMa \cite{behley2018efficient}  & 1.39 \% & 0.0034 [deg/m] & 0.1 s  \\
\hline
3DOF-SLAM \cite{dimitrievski2016robust}  & 1.89 \% & 0.0083 [deg/m] & 0.02 s  \\
\hline
DeepCLR \cite{horn2020deepclr}  & 3.83 \% & 0.0104 [deg/m] & 0.05 s  \\
\hline
D3DLO \cite{adis2021d3dlo}  & 5.40 \% & 0.0154 [deg/m] & 0.1 s \\
\hline
\end{tabular}
\vspace{-5mm}
\end{table}

\subsection{Network correspondence  based methods}

The network-based approaches employ neural networks in the pose estimation module of the LiDAR odometry pipeline. Approaches in the domain of LiDAR odometry using deep learning as the point cloud registration step are relatively recent when compared to the ICP and NDT-based approaches. Nevertheless, there has been a considerable number of works in the deep learning based Visual odometry like \cite{wang2019recurrent} \cite{zou2020learning} \cite{yang2020d3vo}.  

The PointNet paper \cite{qi2017pointnet} (and PointNet++ \cite{qi2017pointnet++}) a popular paper, proposed for point cloud object classification and part segmentation tasks, incorporates a novel network that directly takes in point clouds instead of having to use voxelization like in other approaches. 

Next, an end-to-end method called LO-Net \cite{li2019net} was introduced that takes in LiDAR point cloud data and computes inter-scan 6-DoF relative pose. With it being an end-to-end trainable method, LO-Net learns an effective feature representation. This is facilitated by a new mask-weighted geometric constraint loss. This loss helps the algorithm encash the data's sequential dependencies and dynamics. Here, the position and the orientation are estimated simultaneously.  L3-Net \cite{lu2019l3} has a blend of multiple approaches with mini-PointNet being used for feature extraction and 3DCNNs being used for regularization. Most approaches are supervised learning approaches, but for the first time, the paper DeepLO \cite{cho2019deeplo} introduces supervised and unsupervised frameworks for geometry-aware LiDAR odometry. DeepLO also incorporates vertex and normal maps as network inputs without precision loss.

LodoNet \cite{zheng2020lodonet} adapted the PointNet classification architecture into its rotation and translation estimation modules. But unlike LO-Net, the translation and the rotation modules are separate, resulting in two 3-DoF predictions at the end of odometry. Similar to DCP \cite{wang2019deep} and LeGOLOAM \cite{shan2018lego}, spherical projection of the input LiDAR point clouds is made in approaches like LodoNet, in order to represent the 3-Dimensional data in 2-Dimensions which facilitates utilization of relatively less compute resources. The main constraint with the standard ICP approach to the LiDAR odometry problem is that due to the coupled nature of the odometry regression and the keypoint matching functions, there is a potential issue with training loss non-convergence as delineated in \cite{li2019net}. Whereas in LodoNet, in order to increase the robustness and the effectiveness of network, an MKP selection module based on PointNet \cite{qi2017pointnet} is employed to solve the segmentation problem.
To tackle multi-view problems was a challenge that was yet to be solved for LiDAR odometry applications. An approach along these lines is the 3DRegNet \cite{pais20203dregnet} which can extend for scenarios that involve handling multiple views, not just two like with classical methods. In this approach, in order to classify the point correspondences and regress motion parameters for common reference frame scan alignment, convolutional layers and deep residual layers are incorporated into the neural network. 

The most recent works are D3DLO \cite{adis2021d3dlo}, PWCLO \cite{wang2021pwclo} and OverlapNet \cite{chen2021overlapnet}. Even though D3DLO and DeepCLR \cite{horn2020deepclr} have similar network architectures, D3DLO utilizes 3.56\% of the network parameters. In doing so, it slightly underperforms compared to DeepCLR but manages to reduce the point cloud size by up to 40-50\%. On the other hand, PWCLO, with hierarchical embedding mask optimization, outperforms LodoNet, LOAM, DMLO, and LO-Net in terms of translational and rotational errors on the KITTI odometry sequences.

The network based approaches have improved drastically over the years. Methods like D3DLO \cite{adis2021d3dlo} have matched runtime of LOAM \cite{zhang2014loam} and DeepCLR \cite{horn2020deepclr} has even performed twice as better in terms of runtime.

\section{Performance Comparison}
\label{sec:performance-comparison}
All the LiDAR-only odometry methods as shown in Table \ref{tab:KITTI-training} and Table \ref{tab:KITTI-testing} \cite{geiger2012we} have been compared on the publicly KITTI odometry dataset \cite{geiger2012we}.In the training performance on the KITTI odometry data, Table \ref{tab:KITTI-training}, DeepLO \cite{cho2019deeplo} seems to perform well in most of the sequences compared to the other methods. DeepVCP \cite{lu2019deepvcp} gives the best average performance over all the 11 sequences, compared to others. From Table \ref{tab:KITTI-testing}, the NDT-based LOAM is the best performing method for LiDAR odometry with least rotational and transitional errors. But, many other methods match or even beat the runtime performance metric. ELO \cite{zheng2021efficient} has the best runtime among all the listed ones, with a runtime of 0.005s. Runtime is a very important metric for utilization and deployment into automated driving systems. MULLS \cite{pan2021mulls}, CPFG-SLAM \cite{ji2018cpfg}, 3DOF-SLAM \cite{dimitrievski2016robust}, DeepCLR \cite{horn2020deepclr}, F-LOAM \cite{oelsch2021r} and ELO \cite{zheng2021efficient} outperform LOAM in terms of runtime performance.\\ 
The Efficient LiDAR Odometry: ELO \cite{zheng2021efficient} method with a comparable error performance to LOAM and with a much superior runtime in the test set, also outperforms LOAM in the average performance in training set. ELO seems to be the best approach for the application of real-time LiDAR Odometry for autonomous driving.

\section{Conclusion}
\label{sec:Conclusion}
In this paper, the existing works on LiDAR odometry are surveyed and categorized as point correspondence, distribution correspondence, and network correspondence based methodologies. We also show the evaluations on the KITTI odometry dataset. In the survey, we found that each approach has its advantages and disadvantages. Also, that there are several works that explore the ways to fuse different approaches for better odometry estimation, \emph{e.g.} DCP \cite{wang2019deep} use deep neural networks to generate point correspondences, which achieves promising results. In regards to the future direction of research, fusion-based approaches are suggested for precise LiDAR odometry.

\clearpage

{
\bibliographystyle{IEEEtran}
\bibliography{IEEEfull}

\begin{thebibliography}{10}
\providecommand{\url}[1]{#1}
\csname url@rmstyle\endcsname
\providecommand{\newblock}{\relax}
\providecommand{\bibinfo}[2]{#2}
\providecommand\BIBentrySTDinterwordspacing{\spaceskip=0pt\relax}
\providecommand\BIBentryALTinterwordstretchfactor{4}
\providecommand\BIBentryALTinterwordspacing{\spaceskip=\fontdimen2\font plus
\BIBentryALTinterwordstretchfactor\fontdimen3\font minus
  \fontdimen4\font\relax}
\providecommand\BIBforeignlanguage[2]{{%
\expandafter\ifx\csname l@#1\endcsname\relax
\typeout{** WARNING: IEEEtran.bst: No hyphenation pattern has been}%
\typeout{** loaded for the language `#1'. Using the pattern for}%
\typeout{** the default language instead.}%
\else
\language=\csname l@#1\endcsname
\fi
#2}}

\bibitem{zhu2018gnss}
N.~Zhu, J.~Marais, D.~B{\'e}taille, and M.~Berbineau, ``Gnss position integrity
  in urban environments: A review of literature,'' \emph{IEEE Transactions on
  Intelligent Transportation Systems}, vol.~19, no.~9, pp. 2762--2778, 2018.

\bibitem{tarrio2015realtime}
J.~J. Tarrio and S.~Pedre, ``Realtime edge-based visual odometry for a
  monocular camera,'' in \emph{Proceedings of the IEEE International Conference
  on Computer Vision}, 2015, pp. 702--710.

\bibitem{lu2015robust}
Y.~Lu and D.~Song, ``Robust rgb-d odometry using point and line features,'' in
  \emph{Proceedings of the IEEE International Conference on Computer Vision},
  2015, pp. 3934--3942.

\bibitem{zihao2017event}
A.~Zihao~Zhu, N.~Atanasov, and K.~Daniilidis, ``Event-based visual inertial
  odometry,'' in \emph{Proceedings of the IEEE Conference on Computer Vision
  and Pattern Recognition}, 2017, pp. 5391--5399.

\bibitem{wang2017stereo}
R.~Wang, M.~Schworer, and D.~Cremers, ``Stereo dso: Large-scale direct sparse
  visual odometry with stereo cameras,'' in \emph{Proceedings of the IEEE
  International Conference on Computer Vision}, 2017, pp. 3903--3911.

\bibitem{lianos2018vso}
K.-N. Lianos, J.~L. Schonberger, M.~Pollefeys, and T.~Sattler, ``Vso: Visual
  semantic odometry,'' in \emph{Proceedings of the European conference on
  computer vision (ECCV)}, 2018, pp. 234--250.

\bibitem{yang2018deep}
N.~Yang, R.~Wang, J.~Stuckler, and D.~Cremers, ``Deep virtual stereo odometry:
  Leveraging deep depth prediction for monocular direct sparse odometry,'' in
  \emph{Proceedings of the European Conference on Computer Vision (ECCV)},
  2018, pp. 817--833.

\bibitem{wang2019recurrent}
R.~Wang, S.~M. Pizer, and J.-M. Frahm, ``Recurrent neural network for (un-)
  supervised learning of monocular video visual odometry and depth,'' in
  \emph{Proceedings of the IEEE/CVF Conference on Computer Vision and Pattern
  Recognition}, 2019, pp. 5555--5564.

\bibitem{sheng2019unsupervised}
L.~Sheng, D.~Xu, W.~Ouyang, and X.~Wang, ``Unsupervised collaborative learning
  of keyframe detection and visual odometry towards monocular deep slam,'' in
  \emph{Proceedings of the IEEE/CVF International Conference on Computer
  Vision}, 2019, pp. 4302--4311.

\bibitem{zou2020learning}
Y.~Zou, P.~Ji, Q.-H. Tran, J.-B. Huang, and M.~Chandraker, ``Learning monocular
  visual odometry via self-supervised long-term modeling,'' in \emph{Computer
  Vision--ECCV 2020: 16th European Conference, Glasgow, UK, August 23--28,
  2020, Proceedings, Part XIV 16}.\hskip 1em plus 0.5em minus 0.4em\relax
  Springer, 2020, pp. 710--727.

\bibitem{yang2020d3vo}
N.~Yang, L.~v. Stumberg, R.~Wang, and D.~Cremers, ``D3vo: Deep depth, deep pose
  and deep uncertainty for monocular visual odometry,'' in \emph{Proceedings of
  the IEEE/CVF Conference on Computer Vision and Pattern Recognition}, 2020,
  pp. 1281--1292.

\bibitem{aqel2016review}
M.~O. Aqel, M.~H. Marhaban, M.~I. Saripan, and N.~B. Ismail, ``Review of visual
  odometry: types, approaches, challenges, and applications,''
  \emph{SpringerPlus}, vol.~5, no.~1, pp. 1--26, 2016.

\bibitem{roriz2021automotive}
R.~Roriz, J.~Cabral, and T.~Gomes, ``Automotive lidar technology: A survey,''
  \emph{IEEE Transactions on Intelligent Transportation Systems}, 2021.

\bibitem{li2020dmlo}
Z.~Li and N.~Wang, ``Dmlo: Deep matching lidar odometry,'' in \emph{2020
  IEEE/RSJ International Conference on Intelligent Robots and Systems
  (IROS)}.\hskip 1em plus 0.5em minus 0.4em\relax IEEE, 2020, pp. 6010--6017.

\bibitem{shan2018lego}
T.~Shan and B.~Englot, ``Lego-loam: Lightweight and ground-optimized lidar
  odometry and mapping on variable terrain,'' in \emph{2018 IEEE/RSJ
  International Conference on Intelligent Robots and Systems (IROS)}.\hskip 1em
  plus 0.5em minus 0.4em\relax IEEE, 2018, pp. 4758--4765.

\bibitem{zheng2020lodonet}
C.~Zheng, Y.~Lyu, M.~Li, and Z.~Zhang, ``Lodonet: A deep neural network with 2d
  keypoint matching for 3d lidar odometry estimation,'' in \emph{Proceedings of
  the 28th ACM International Conference on Multimedia}, 2020, pp. 2391--2399.

\bibitem{lowe2004distinctive}
D.~G. Lowe, ``Distinctive image features from scale-invariant keypoints,''
  \emph{International journal of computer vision}, vol.~60, no.~2, pp. 91--110,
  2004.

\bibitem{bay2006surf}
H.~Bay, T.~Tuytelaars, and L.~Van~Gool, ``Surf: Speeded up robust features,''
  in \emph{European conference on computer vision}.\hskip 1em plus 0.5em minus
  0.4em\relax Springer, 2006, pp. 404--417.

\bibitem{rublee2011orb}
E.~Rublee, V.~Rabaud, K.~Konolige, and G.~Bradski, ``Orb: An efficient
  alternative to sift or surf,'' in \emph{2011 International conference on
  computer vision}.\hskip 1em plus 0.5em minus 0.4em\relax Ieee, 2011, pp.
  2564--2571.

\bibitem{wang2019deep}
Y.~Wang and J.~M. Solomon, ``Deep closest point: Learning representations for
  point cloud registration,'' in \emph{Proceedings of the IEEE/CVF
  International Conference on Computer Vision}, 2019, pp. 3523--3532.

\bibitem{biber2003normal}
P.~Biber and W.~Stra{\ss}er, ``The normal distributions transform: A new
  approach to laser scan matching,'' in \emph{Proceedings 2003 IEEE/RSJ
  International Conference on Intelligent Robots and Systems (IROS 2003)(Cat.
  No. 03CH37453)}, vol.~3.\hskip 1em plus 0.5em minus 0.4em\relax IEEE, 2003,
  pp. 2743--2748.

\bibitem{chen2021overlapnet}
X.~Chen, T.~L{\"a}be, A.~Milioto, T.~R{\"o}hling, O.~Vysotska, A.~Haag,
  J.~Behley, and C.~Stachniss, ``Overlapnet: Loop closing for lidar-based
  slam,'' \emph{arXiv preprint arXiv:2105.11344}, 2021.

\bibitem{zhang1992iterative}
Z.~Zhang, ``Iterative point matching for registration of free-form curves,''
  Ph.D. dissertation, Inria, 1992.

\bibitem{fischler1981random}
M.~A. Fischler and R.~C. Bolles, ``Random sample consensus: a paradigm for
  model fitting with applications to image analysis and automated
  cartography,'' \emph{Communications of the ACM}, vol.~24, no.~6, pp.
  381--395, 1981.

\bibitem{wang2020lidar}
Y.~Wang, Z.~Sun, C.-Z. Xu, S.~E. Sarma, J.~Yang, and H.~Kong, ``Lidar iris for
  loop-closure detection,'' in \emph{2020 IEEE/RSJ International Conference on
  Intelligent Robots and Systems (IROS)}.\hskip 1em plus 0.5em minus
  0.4em\relax IEEE, 2020, pp. 5769--5775.

\bibitem{hess2016real}
W.~Hess, D.~Kohler, H.~Rapp, and D.~Andor, ``Real-time loop closure in 2d lidar
  slam,'' in \emph{2016 IEEE International Conference on Robotics and
  Automation (ICRA)}.\hskip 1em plus 0.5em minus 0.4em\relax IEEE, 2016, pp.
  1271--1278.

\bibitem{zhang2014loam}
J.~Zhang and S.~Singh, ``Loam: Lidar odometry and mapping in real-time.'' in
  \emph{Robotics: Science and Systems}, vol.~2, no.~9, 2014.

\bibitem{chen2019suma++}
X.~Chen, A.~Milioto, E.~Palazzolo, P.~Giguere, J.~Behley, and C.~Stachniss,
  ``Suma++: Efficient lidar-based semantic slam,'' in \emph{2019 IEEE/RSJ
  International Conference on Intelligent Robots and Systems (IROS)}.\hskip 1em
  plus 0.5em minus 0.4em\relax IEEE, 2019, pp. 4530--4537.

\bibitem{kanatani1994analysis}
K.-i. Kanatani, ``Analysis of 3-d rotation fitting,'' \emph{IEEE Transactions
  on pattern analysis and machine intelligence}, vol.~16, no.~5, pp. 543--549,
  1994.

\bibitem{lee2020robust}
S.~Lee, C.~Kim, S.~Cho, S.~Myoungho, and K.~Jo, ``Robust 3-dimension point
  cloud mapping in dynamic environment using point-wise static
  probability-based ndt scan-matching,'' \emph{IEEE Access}, vol.~8, pp.
  175\,563--175\,575, 2020.

\bibitem{lu2019deepvcp}
W.~Lu, G.~Wan, Y.~Zhou, X.~Fu, P.~Yuan, and S.~Song, ``Deepvcp: An end-to-end
  deep neural network for point cloud registration,'' in \emph{Proceedings of
  the IEEE/CVF International Conference on Computer Vision}, 2019, pp. 12--21.

\bibitem{zheng2021efficient}
X.~Zheng and J.~Zhu, ``Efficient lidar odometry for autonomous driving,''
  \emph{arXiv preprint arXiv:2104.10879}, 2021.

\bibitem{thomas2019delio}
Q.~M. Thomas, O.~Wasenm{\"u}ller, and D.~Stricker, ``Delio: Decoupled lidar
  odometry,'' in \emph{2019 IEEE Intelligent Vehicles Symposium (IV)}.\hskip
  1em plus 0.5em minus 0.4em\relax IEEE, 2019, pp. 1549--1556.

\bibitem{oelsch2021r}
M.~Oelsch, M.~Karimi, and E.~Steinbach, ``R-loam: Improving lidar odometry and
  mapping with point-to-mesh features of a known 3d reference object,''
  \emph{IEEE Robotics and Automation Letters}, vol.~6, no.~2, pp. 2068--2075,
  2021.

\bibitem{li2019net}
Q.~Li, S.~Chen, C.~Wang, X.~Li, C.~Wen, M.~Cheng, and J.~Li, ``Lo-net: Deep
  real-time lidar odometry,'' in \emph{Proceedings of the IEEE/CVF Conference
  on Computer Vision and Pattern Recognition}, 2019, pp. 8473--8482.

\bibitem{zhou2018voxelnet}
Y.~Zhou and O.~Tuzel, ``Voxelnet: End-to-end learning for point cloud based 3d
  object detection,'' in \emph{Proceedings of the IEEE conference on computer
  vision and pattern recognition}, 2018, pp. 4490--4499.

\bibitem{qi2017pointnet}
C.~R. Qi, H.~Su, K.~Mo, and L.~J. Guibas, ``Pointnet: Deep learning on point
  sets for 3d classification and segmentation,'' in \emph{Proceedings of the
  IEEE conference on computer vision and pattern recognition}, 2017, pp.
  652--660.

\bibitem{chetverikov2002trimmed}
D.~Chetverikov, D.~Svirko, D.~Stepanov, and P.~Krsek, ``The trimmed iterative
  closest point algorithm,'' in \emph{Object recognition supported by user
  interaction for service robots}, vol.~3.\hskip 1em plus 0.5em minus
  0.4em\relax IEEE, 2002, pp. 545--548.

\bibitem{segal2009generalized}
A.~Segal, D.~Haehnel, and S.~Thrun, ``Generalized-icp.'' in \emph{Robotics:
  science and systems}, vol.~2, no.~4.\hskip 1em plus 0.5em minus 0.4em\relax
  Seattle, WA, 2009, p. 435.

\bibitem{bouaziz2013sparse}
S.~Bouaziz, A.~Tagliasacchi, and M.~Pauly, ``Sparse iterative closest point,''
  in \emph{Computer graphics forum}, vol.~32, no.~5.\hskip 1em plus 0.5em minus
  0.4em\relax Wiley Online Library, 2013, pp. 113--123.

\bibitem{pavlov2018aa}
A.~L. Pavlov, G.~W. Ovchinnikov, D.~Y. Derbyshev, D.~Tsetserukou, and I.~V.
  Oseledets, ``Aa-icp: Iterative closest point with anderson acceleration,'' in
  \emph{2018 IEEE International Conference on Robotics and Automation
  (ICRA)}.\hskip 1em plus 0.5em minus 0.4em\relax IEEE, 2018, pp. 3407--3412.

\bibitem{parkison2018semantic}
S.~A. Parkison, L.~Gan, M.~G. Jadidi, and R.~M. Eustice, ``Semantic iterative
  closest point through expectation-maximization.'' in \emph{BMVC}, 2018, p.
  280.

\bibitem{choy2020deep}
C.~Choy, W.~Dong, and V.~Koltun, ``Deep global registration,'' in
  \emph{Proceedings of the IEEE/CVF conference on computer vision and pattern
  recognition}, 2020, pp. 2514--2523.

\bibitem{grant2012point}
D.~Grant, J.~Bethel, and M.~Crawford, ``Point-to-plane registration of
  terrestrial laser scans,'' \emph{ISPRS Journal of Photogrammetry and Remote
  Sensing}, vol.~72, pp. 16--26, 2012.

\bibitem{chen1992object}
Y.~Chen and G.~Medioni, ``Object modelling by registration of multiple range
  images,'' \emph{Image and vision computing}, vol.~10, no.~3, pp. 145--155,
  1992.

\bibitem{deschaud2018imls}
J.-E. Deschaud, ``Imls-slam: scan-to-model matching based on 3d data,'' in
  \emph{2018 IEEE International Conference on Robotics and Automation
  (ICRA)}.\hskip 1em plus 0.5em minus 0.4em\relax IEEE, 2018, pp. 2480--2485.

\bibitem{kovalenko2019sensor}
D.~Kovalenko, M.~Korobkin, and A.~Minin, ``Sensor aware lidar odometry,'' in
  \emph{2019 European Conference on Mobile Robots (ECMR)}.\hskip 1em plus 0.5em
  minus 0.4em\relax IEEE, 2019, pp. 1--6.

\bibitem{milioto2019rangenet++}
A.~Milioto, I.~Vizzo, J.~Behley, and C.~Stachniss, ``Rangenet++: Fast and
  accurate lidar semantic segmentation,'' in \emph{2019 IEEE/RSJ International
  Conference on Intelligent Robots and Systems (IROS)}.\hskip 1em plus 0.5em
  minus 0.4em\relax IEEE, 2019, pp. 4213--4220.

\bibitem{behley2018efficient}
J.~Behley and C.~Stachniss, ``Efficient surfel-based slam using 3d laser range
  data in urban environments.'' in \emph{Robotics: Science and Systems}, vol.
  2018, 2018.

\bibitem{rufus2020srom}
N.~Rufus, U.~K.~R. Nair, A.~S.~B. Kumar, V.~Madiraju, and K.~M. Krishna,
  ``Srom: Simple real-time odometry and mapping using lidar data for autonomous
  vehicles,'' in \emph{2020 IEEE Intelligent Vehicles Symposium (IV)}.\hskip
  1em plus 0.5em minus 0.4em\relax IEEE, 2020, pp. 1867--1872.

\bibitem{zhu2021cylindrical}
X.~Zhu, H.~Zhou, T.~Wang, F.~Hong, Y.~Ma, W.~Li, H.~Li, and D.~Lin,
  ``Cylindrical and asymmetrical 3d convolution networks for lidar
  segmentation,'' in \emph{Proceedings of the IEEE/CVF Conference on Computer
  Vision and Pattern Recognition}, 2021, pp. 9939--9948.

\bibitem{magnusson20083d}
M.~Magnusson, A.~N{\"u}chter, C.~L{\"o}rken, A.~J. Lilienthal, and
  J.~Hertzberg, ``3d mapping the kvarntorp mine: a rield experiment for
  evaluation of 3d scan matching algorithms,'' in \emph{IEEE/RSJ International
  Conference on Intelligent Robots and Systems (IROS), Workshop" 3D Mapping",
  Nice, France, September 2008}, 2008.

\bibitem{hong2017probabilistic}
H.~Hong and B.~H. Lee, ``Probabilistic normal distributions transform
  representation for accurate 3d point cloud registration,'' in \emph{2017
  IEEE/RSJ International Conference on Intelligent Robots and Systems
  (IROS)}.\hskip 1em plus 0.5em minus 0.4em\relax IEEE, 2017, pp. 3333--3338.

\bibitem{akai2017robust}
N.~Akai, L.~Y. Morales, E.~Takeuchi, Y.~Yoshihara, and Y.~Ninomiya, ``Robust
  localization using 3d ndt scan matching with experimentally determined
  uncertainty and road marker matching,'' in \emph{2017 IEEE Intelligent
  Vehicles Symposium (IV)}.\hskip 1em plus 0.5em minus 0.4em\relax IEEE, 2017,
  pp. 1356--1363.

\bibitem{chen2020sloam}
S.~W. Chen, G.~V. Nardari, E.~S. Lee, C.~Qu, X.~Liu, R.~A.~F. Romero, and
  V.~Kumar, ``Sloam: Semantic lidar odometry and mapping for forest
  inventory,'' \emph{IEEE Robotics and Automation Letters}, vol.~5, no.~2, pp.
  612--619, 2020.

\bibitem{wang2020intensity}
H.~Wang, C.~Wang, and L.~Xie, ``Intensity scan context: Coding intensity and
  geometry relations for loop closure detection,'' in \emph{2020 IEEE
  International Conference on Robotics and Automation (ICRA)}.\hskip 1em plus
  0.5em minus 0.4em\relax IEEE, 2020, pp. 2095--2101.

\bibitem{wang2021f}
H.~Wang, C.~Wang, C.-L. Chen, and L.~Xie, ``F-loam: Fast lidar odometry and
  mapping,'' \emph{arXiv preprint arXiv:2107.00822}, 2021.

\bibitem{cho2019deeplo}
Y.~Cho, G.~Kim, and A.~Kim, ``Deeplo: Geometry-aware deep lidar odometry,''
  \emph{arXiv preprint arXiv:1902.10562}, 2019.

\bibitem{pan2021mulls}
Y.~Pan, P.~Xiao, Y.~He, Z.~Shao, and Z.~Li, ``Mulls: Versatile lidar slam via
  multi-metric linear least square,'' \emph{arXiv preprint arXiv:2102.03771},
  2021.

\bibitem{neuhaus2018mc2slam}
F.~Neuhaus, T.~Ko{\ss}, R.~Kohnen, and D.~Paulus, ``Mc2slam: Real-time inertial
  lidar odometry using two-scan motion compensation,'' in \emph{German
  Conference on Pattern Recognition}.\hskip 1em plus 0.5em minus 0.4em\relax
  Springer, 2018, pp. 60--72.

\bibitem{chen2020psf}
G.~Chen, B.~Wang, X.~Wang, H.~Deng, B.~Wang, and S.~Zhang, ``Psf-lo:
  Parameterized semantic features based lidar odometry,'' \emph{arXiv preprint
  arXiv:2010.13355}, 2020.

\bibitem{yin2020cae}
D.~Yin, Q.~Zhang, J.~Liu, X.~Liang, Y.~Wang, J.~Maanp{\"a}{\"a}, H.~Ma,
  J.~Hyypp{\"a}, and R.~Chen, ``Cae-lo: Lidar odometry leveraging fully
  unsupervised convolutional auto-encoder for interest point detection and
  feature description,'' \emph{arXiv preprint arXiv:2001.01354}, 2020.

\bibitem{ji2018cpfg}
K.~Ji, H.~Chen, H.~Di, J.~Gong, G.~Xiong, J.~Qi, and T.~Yi, ``Cpfg-slam: A
  robust simultaneous localization and mapping based on lidar in off-road
  environment,'' in \emph{2018 IEEE Intelligent Vehicles Symposium (IV)}.\hskip
  1em plus 0.5em minus 0.4em\relax IEEE, 2018, pp. 650--655.

\bibitem{chen2021moving}
X.~Chen, S.~Li, B.~Mersch, L.~Wiesmann, J.~Gall, J.~Behley, and C.~Stachniss,
  ``Moving object segmentation in 3d lidar data: A learning-based approach
  exploiting sequential data,'' \emph{arXiv preprint arXiv:2105.08971}, 2021.

\bibitem{yoon2021unsupervised}
D.~J. Yoon, H.~Zhang, M.~Gridseth, H.~Thomas, and T.~D. Barfoot, ``Unsupervised
  learning of lidar features for use ina probabilistic trajectory estimator,''
  \emph{IEEE Robotics and Automation Letters}, vol.~6, no.~2, pp. 2130--2138,
  2021.

\bibitem{tang2019white}
T.~Y. Tang, D.~J. Yoon, and T.~D. Barfoot, ``A white-noise-on-jerk motion prior
  for continuous-time trajectory estimation on se (3),'' \emph{IEEE Robotics
  and Automation Letters}, vol.~4, no.~2, pp. 594--601, 2019.

\bibitem{dimitrievski2016robust}
M.~Dimitrievski, D.~Van~Hamme, P.~Veelaert, and W.~Philips, ``Robust matching
  of occupancy maps for odometry in autonomous vehicles,'' in \emph{11th Joint
  Conference on Computer Vision, Imaging and Computer Graphics Theory and
  Applications (VISIGRAPP 2016)}, vol.~3, 2016, pp. 626--633.

\bibitem{horn2020deepclr}
M.~Horn, N.~Engel, V.~Belagiannis, M.~Buchholz, and K.~Dietmayer, ``Deepclr:
  Correspondence-less architecture for deep end-to-end point cloud
  registration,'' in \emph{2020 IEEE 23rd International Conference on
  Intelligent Transportation Systems (ITSC)}.\hskip 1em plus 0.5em minus
  0.4em\relax IEEE, 2020, pp. 1--7.

\bibitem{adis2021d3dlo}
P.~Adis, N.~Horst, and M.~Wien, ``D3dlo: Deep 3d lidar odometry,'' \emph{arXiv
  preprint arXiv:2101.12242}, 2021.

\bibitem{qi2017pointnet++}
C.~R. Qi, L.~Yi, H.~Su, and L.~J. Guibas, ``Pointnet++: Deep hierarchical
  feature learning on point sets in a metric space,'' \emph{arXiv preprint
  arXiv:1706.02413}, 2017.

\bibitem{lu2019l3}
W.~Lu, Y.~Zhou, G.~Wan, S.~Hou, and S.~Song, ``L3-net: Towards learning based
  lidar localization for autonomous driving,'' in \emph{Proceedings of the
  IEEE/CVF Conference on Computer Vision and Pattern Recognition}, 2019, pp.
  6389--6398.

\bibitem{pais20203dregnet}
G.~D. Pais, S.~Ramalingam, V.~M. Govindu, J.~C. Nascimento, R.~Chellappa, and
  P.~Miraldo, ``3dregnet: A deep neural network for 3d point registration,'' in
  \emph{Proceedings of the IEEE/CVF conference on computer vision and pattern
  recognition}, 2020, pp. 7193--7203.

\bibitem{wang2021pwclo}
G.~Wang, X.~Wu, Z.~Liu, and H.~Wang, ``Pwclo-net: Deep lidar odometry in 3d
  point clouds using hierarchical embedding mask optimization,'' in
  \emph{Proceedings of the IEEE/CVF Conference on Computer Vision and Pattern
  Recognition}, 2021, pp. 15\,910--15\,919.

\bibitem{geiger2012we}
A.~Geiger, P.~Lenz, and R.~Urtasun, ``Are we ready for autonomous driving? the
  kitti vision benchmark suite,'' in \emph{2012 IEEE conference on computer
  vision and pattern recognition}.\hskip 1em plus 0.5em minus 0.4em\relax IEEE,
  2012, pp. 3354--3361.

\end{thebibliography}
}

\end{document}